\pdfoutput=1

\documentclass[11pt]{article}

\usepackage{acl}

\usepackage{times}
\usepackage{latexsym}
\usepackage{spverbatim}
\usepackage{changepage}   
\usepackage{graphicx}

\usepackage{booktabs}

\usepackage[T1]{fontenc}

\usepackage[utf8]{inputenc}

\usepackage{microtype}

\newcommand{\languagefamilycountacrossalldatasets}{32}
\newcommand{\languagecountacrossalldatasets}{363}
\newcommand{\languagecountinbloomvistafterfiltering}{351}
\newcommand{\languagecountinbloomspeech}{56}

\title{Bloom Library: Multimodal Datasets in 300+ Languages for a Variety of Downstream Tasks}

\author{Colin Leong\textsuperscript{*}, Joshua Nemecek\textsuperscript{\textdagger}, 
        Jacob Mansdorfer\textsuperscript{\textsection}, Anna Filighera\textsuperscript{\textparagraph}, \\  \textbf{Abraham Owodunni}\textsuperscript{\textbardbl} \and \textbf{Daniel Whitenack}\textsuperscript{\textdagger} \\
        \textsuperscript{*}University of Dayton Research Institute,\textsuperscript{\textdagger}SIL International,\textsuperscript{\textsection}Independent Contractor, \\ \textsuperscript{\textparagraph}TU Darmstadt \and \textsuperscript{\textbardbl}Masakhane \\
        \texttt{cleong1@udayton.edu}, \texttt{joshua\_nemecek@sil.org} \\
        \texttt{jacob.mansdorfer@gmail.com}, \texttt{anna.filighera@kom.tu-darmstadt.de} \\
        \texttt{owodunniabraham@gmail.com}, \texttt{dan\_whitenack@sil.org}
        }

\begin{document}
\maketitle
\begin{abstract}
We present Bloom Library, a linguistically diverse set of multimodal and multilingual datasets for language modeling, image captioning, visual storytelling, and speech synthesis/recognition. These datasets represent either the most, or among the most, multilingual datasets for each of the included downstream tasks. In total, the initial release of the Bloom Library datasets covers \languagecountacrossalldatasets{} languages across \languagefamilycountacrossalldatasets{} language families. We train downstream task models for various languages represented in the data, showing the viability of the data for future work in low-resource, multimodal NLP and establishing the first known baselines for these downstream tasks in certain languages (e.g., Bisu [bzi], with an estimated population of 700 users). Some of these first-of-their-kind baselines are comparable to state-of-the-art performance for higher-resourced languages. The Bloom Library datasets are released under Creative Commons licenses on the Hugging Face datasets hub to catalyze more linguistically diverse research in the included downstream tasks.
\end{abstract}

\section{Introduction}

Only a negligible fraction of the 7100+ living languages~\cite{david_m__eberhard_ethnologue_2021} have sufficient, publicly available text, audio, and image data to train state-of-the-art language/speech models and/or models for downstream tasks like Named Entity Recognition (NER) or image captioning. This data scarcity results in systematic inequalities in the performance of NLP tasks across the world's languages~\cite{Blasi2021SystematicII}. Indigenous language ecologies also represent profoundly different understandings of the nature and function of language~\cite{bird-2022-local, bird-2020-decolonising}, which might prioritize orality or translanguaging~\cite{quakenbush2018language}, for example, above a single, written mode of communication in all domains.  

The Bloom Library\footnote{\url{https://bloomlibrary.org/}} is a web-based platform that is attempting to facilitate an increase in the amount of multimodal materials available to communities speaking non-dominant languages. The Bloom Library holds over 12,400 books in 545 languages (at the time this paper is published), covering subjects including agriculture, business, culture, math, science, religion, and health. Many of these books include images aligned with text, and 1,600+ of the books have corresponding audio recordings (called "talking books"). Language communities can create new books, create audio recordings, download existing books, and translate existing books using the open-source "Bloom" software\footnote{\url{https://github.com/BloomBooks/BloomDesktop}}.  

To boost language diversity and indigenous perspectives in the NLP research community, we present multimodal datasets post-processed out of the Bloom Library. We anticipate that more task-specific datasets will be created from the Bloom Library. However, as a starting point, we are presenting the following datasets: (1) \texttt{bloom-lm} for language modeling in \languagecountinbloomvistafterfiltering{} languages; (2) \texttt{bloom-captioning} for image-to-text or text-to-image tasks in \languagecountinbloomvistafterfiltering{} languages; (3) \texttt{bloom-vist} for visual storytelling in \languagecountinbloomvistafterfiltering{} languages; and (4) \texttt{bloom-speech} for speech-to-text and text-to-speech tasks in \languagecountinbloomspeech{} languages.

The languages in these datasets correspond to \languagefamilycountacrossalldatasets{} language families, and many of the included languages are in extremely low-resource settings. Further, to the authors' knowledge, \texttt{bloom-vist} represents the first (and certainly most) multilingual visual storytelling dataset, and \texttt{bloom-speech} includes more languages in the following language families than any other aligned speech dataset (number of languages in parenthesis): Austronesian (8), Mayan (6), Niger-Congo (7), Sepik (2), Tequistlatecan (2), and Trans-New Guinea (3).

To assess the difficulty of language modeling, image captioning, and automatic speech recognition (ASR) with the Bloom Library datasets, we trained baseline models on each of these tasks. For certain languages, the Bloom Library datasets facilitate the first known baselines with comparable to state-of-the-art performance for higher-resourced languages. We acheive a BLEU score on image captioning of above 10.0 for 10 languages using only data from \texttt{bloom-captioning}. For ASR, we demonstrate a Word Error Rate (WER) below 0.5 for 18 languages and a Character Error Rate (CER) below 0.2 for 21 languages.

\section{Related Work}

In terms of language coverage, various multilingual and single modality datasets have emerged recently. These include, by way of example, the JHU Bible Corpus~\cite{mccarthy-etal-2020-johns}, the CMU Wilderness Multilingual Speech dataset~\cite{8683536}, Common Voice 9~\cite{commonvoice}, Multilingual BABEL~\cite{babel-ldc}, and MASSIVE~\cite{MASSIVE}. The number of languages in these datasets is impressive. However, many are limited in domain (e.g., only including Bible data), accessibility, licensing, or modality (e.g., only focusing on text or read speech). These datasets are also primarily rooted in content from large, dominant languages, like English, and are translated or adapted to other fairly large languages. Bloom Library data, in contrast, originates from local language communities,\footnote{On the use of the term “local” languages, we followed the terminology used in \citet{bird-2022-local} and related works, which defines the term along the lines of “small, primarily-oral languages, often Indigenous or endangered, including the original and emerging languages of Africa, Asia, Australia, the Americas, the Pacific, and the minority languages of Europe.”} which are producing Bloom Books to fit their own local language ecology and perspectives. As a result, the data presented here covers languages, language families, and topics that are not covered by any other aligned and prepared datasets.

In terms of modality, the research community is presenting an increasing number of intriguing multimodal datasets. These include, by way of example, Pano-AVQA~\cite{Yun2021PanoAVQA}, which facilitates question answering regarding various objects, sounds, and their associations in videos, and VIST~\cite{huang2016visual}, which facilitates sequential vision-to-language tasks. However, recent multimodal datasets are overwhelmingly monolingual.

Datasets representing both multiple modalities and many languages include Multi30k, which is one of the few multimodal, multilingual datasets in existence, with \textasciitilde30k images and corresponding text descriptions in several languages including English, German~\cite{W16-3210}, French~\cite{elliott-EtAl:2017:WMT}, and Czech~\cite{barrault2018findings}. One listing can be found in \citet{Kdr2019LearningVG}, which provides a helpful (and comprehensive) table of multilingual, multimodal resources, dividing them into two categories: (i) "translation" (with captions translated into another language); and (ii) "description" (with annotations independently created for each language). The table reveals that Multi30k was, at the time, the largest translation dataset available in terms of image count, at approximately 31k images and 31k sentences covering 4 languages. 

The Bloom Library datasets fit into the "description" category of \citet{Kdr2019LearningVG}. However, with over 90k+ images and 110k+ captions covering \languagecountinbloomvistafterfiltering{} languages and additional speech data in \languagecountinbloomspeech{} languages, Bloom Library represents a massive increase in language and modality coverage (up to two orders of magnitude wider than previous multilingual, multimodal datasets). Further, the existing datasets referenced by \citet{Kdr2019LearningVG} focus on large languages in high-resource settings, with no representation of local languages in low resource settings. In contrast, our datasets include languages in extremely low resource and non-dominant settings like Bisu [bzi] and Kagayanen [cgc], with estimated populations of 700 and 30,000 users, respectively.


\section{Constructing the Datasets}

The authors worked directly with the Bloom Library developers to gain access to and understand the raw data behind the Bloom Library website. We parse, clean, deduplicate, and publicly release this data for research use on the Hugging Face Hub\footnote{\url{https://www.ai.sil.org/bloom}}\footnote{\url{https://huggingface.co/sil-ai}} in formats compatible with the Hugging Face \texttt{datasets} Python package.\footnote{\url{https://huggingface.co/docs/datasets/index}} 

\texttt{bloom-lm}, \texttt{bloom-captioning}, and \texttt{bloom-vist} are created using one data pipeline starting with \texttt{bloom-vist}, because each of these datasets use some or all of the images and corresponding text within the Bloom Library. A separate data pipeline is used for \texttt{bloom-speech} to process only "talking books."  

\subsection{\texttt{bloom-vist}}

The Bloom Library books offer the rare possibility of leveraging sequential images for language understanding across many languages. Thus, we first process the Bloom Library data into a format consistent with the VIST task published by \citet{huang2016visual}. VIST is a dataset made by creating collections of sequential image-caption pairs which form short “stories”, collaboratively setup by researchers at Google Research, CMU, and JHU, we structure our data to match this We hope this release of VIST-formatted data from Bloom Library catalyzes techniques in both multilingual and multimodal storytelling. 

\begin{figure*}[ht]
  \centering
  \includegraphics[width=\textwidth]{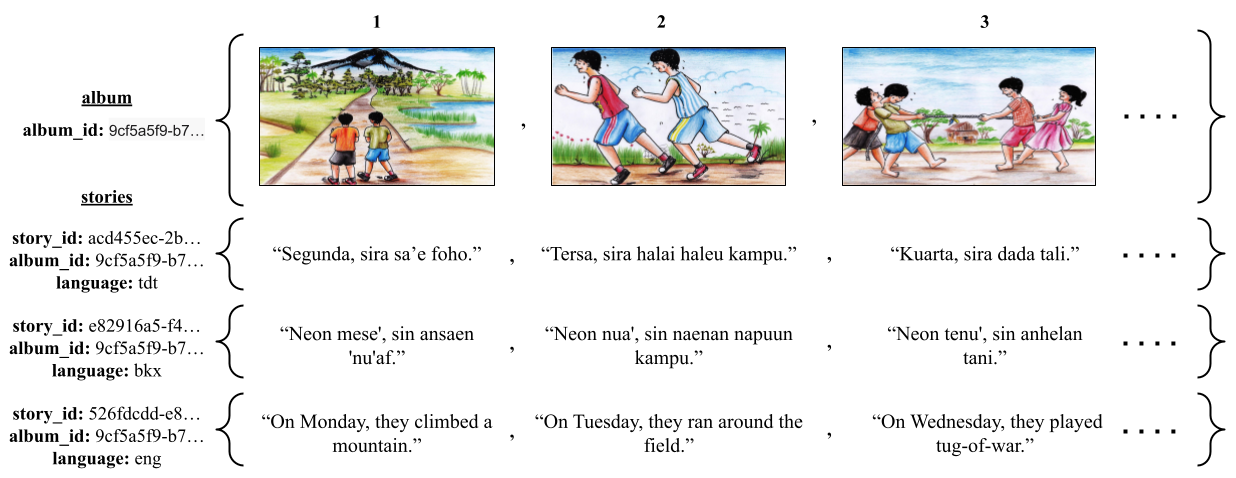}
  \caption{Several examples from the \texttt{bloom-vist} dataset. Three stories (sets of captions) associated with the same album (set of images) are shown, in this case corresponding with the book titled "Mepu"\footnote{\url{https://bloomlibrary.org/book/9BxRXIc74G}} }
  \label{fig:vist}
\end{figure*}

The raw Bloom Library data we received from the Bloom Library team consisted of a folder of files for each "book," which corresponds to one of the pages on the Bloom Library website. The relevant files in this folder include: (1) \texttt{meta.json}, containing important metadata such as the book's translation lineage, alternative titles, copyright, etc.; (2) an \texttt{*.htm} file containing the actual data, particularly text and image links for each page of the book; (3) in certain cases, a number of image files of various types including *.jpg and *.png; and (4) in certain cases (for talking books), a number of \texttt{*.mp3} audio files. In order to construct the sequential VIST-type data, we parse the \texttt{*.htm} file with BeautifulSoup\footnote{\url{https://www.crummy.com/software/BeautifulSoup/}} to associate images files with captions and sequence these according to the sequence of pages in the book. We use \texttt{meta.json} to pull out relevant metadata (book title, topics, etc.) and to filter out any books not released under a Creative Commons license. 

Figure~\ref{fig:vist} includes some example data included in \texttt{bloom-vist} by way of example. The dataset includes "albums," which are ordered sequences of images. An album may be associated with multiple "stories," where each story is an ordered sequence of text captions. 

Once in the appropriate format, we take various steps to clean up and filter the data. We check for, among other things, irreconcilable inconsistencies in metadata (like  conflicting titles or book IDs), duplicate books, duplicate stories, duplicate albums, and similar or identical image-caption pairs. To account for image size or brightness variations during deduplication, we utilize a perceptual hash\footnote{\url{https://github.com/JohannesBuchner/imagehash}} to identify albums sharing at least 80\% of images. We also filter out stories where the writing system script (e.g., Latn or Thai) does not match the majority writing system script used for that language. Of the 14,095 stories in the raw data, 2,547 were duplicates and 155 are filtered due to script mismatch. 

Finally, we follow \citet{kreutzer-etal-2022-quality} and conduct a manual inspection for every language, rejecting any with obvious quality issues "at a glance." As in that work, some of the authors\footnote{Colin Leong: Native English and L2 Mandarin Chinese, and Anna Filighera: Native German and L2 French.} conducted manual inspections on languages they were familiar with (e.g. Mandarin, German), but also languages they had no familiarity with. These checks provide a "floor" on data quality, allowing the detection of extremely low-quality data that is quite obviously wrong even at a glance even by those who do not speak the language.

For example, in this manual review, we detected a number of books having captions in the wrong language (e.g. "English" text in Devanagari or Arabic script) or obvious "test" stories containing the verbatim phrases \textit{"text in a block"} or the English text \textit{"THIS IS ALSO IN FALI."} in a book marked as being in the Fali language. Manual inspection was conducted on at least 50 random stories per language - or fewer if there were fewer stories in a language overall. 85 stories did not pass this manual inspection, some of which were also filtered out by the other quality checks.

Stories which failed any of the checks above are marked as "quarantined" in the JSON file. Downstream data loading scripts can then filter these when loading the data.




After all filtering and "quarantining" of items in the JSON, we are left with 11,407 stories containing a total of 112,080 image/caption pairs in our dataset listed on HuggingFace. The \texttt{bloom-vist} dataset is listed in the Hugging Face Hub as \textit{bloom-vist}.\footnote{\url{https://huggingface.co/datasets/sil-ai/bloom-vist}}

\subsection{\texttt{bloom-captioning}}

Building off of the data produced for \texttt{bloom-vist}, we further process the VIST JSON data into a format useful for image-captioning (and other text-to-image or image-to-text) tasks. More specifically, we post-process the images and corresponding captions into a format consistent with that of the Red Caps dataset~\cite{desai2021redcaps}.\footnote{\url{https://huggingface.co/datasets/red_caps}}. This format includes (for each sample in each language) the text caption,  the album ID, the image ID, and a static URL to a publicly accessible image file. Additionally and in contrast to the Red Caps data, we include language metadata for each set of captions including a normalized ISO639-3 language code as well as the original language code parsed from the Bloom Library files.

In order to prevent data leakage, all image-caption pairs from one story are put into the same split. If there are fewer than 50 stories for a language, we only provide a test split for that language. If there are more than 50 stories for a language, the validation split receives 20\% of the next 250 stories and the train split receives 80\%. Any stories above 300 are split between train (90\%) and test(10\%).

In \texttt{bloom-captioning}, a total of 112,080 image-caption pairs are included, with 157 languages have training splits. For the other languages, the data may be used for testing, or for zero-shot experiments. The \texttt{bloom-captioning} dataset is listed in the Hugging Face Hub as \textit{bloom-captioning}.\footnote{\url{https://huggingface.co/datasets/sil-ai/bloom-captioning}}

\subsection{\texttt{bloom-lm}}

Building off of the data produced for \texttt{bloom-captioning}, we further process the captioning data into a format useful for language modeling (and other written language NLP) tasks. More specifically, we concatenate all of the captions from each story into a single story text. This results in an array of texts per language. For each language, this array is randomized and split with 80\% going into the train set and the remaining 20\% split evenly between test and validation.  

The \texttt{bloom-captioning} dataset is listed in the Hugging Face Hub as \textit{bloom-lm}.\footnote{\url{https://huggingface.co/datasets/sil-ai/bloom-lm}}

\subsection{\texttt{bloom-speech}}

We construct the \texttt{bloom-speech} dataset separately from the data pipeline described above for image and caption related data. Bloom "Talking Books" also have an \texttt{*.htm} file associated with them. We parse the HTML tags within this file that contains the information about the paired audio file and text. The language for each audio file was present in either the same tag, or in one of the parent tags (for example, a tag for the whole page of a book with individual sentences underneath).  As there is some re-use of audio between books, only one audio segment for each language was downloaded per unique text string. 

\defcitealias{scriptsource.org}{www.scriptsource.org}
For all files that were successfully parsed and downloaded, additional checks were performed to ensure that these files matched the tagged language. First, we evaluate the writing system scripts for each language using a mapping of unicode character ranges to script types.\footnote{As indicated by \citetalias{scriptsource.org}} Text strings that did not match the script typed used by the majority of records for that language were thrown out. For some languages, this filtering may exclude known alternate script types. However, based on our manual review, the filtering primarily removed records tagged with incorrect language tags. Additionally, we applied FastText language identification~\cite{fasttextclass} to each text string. If the book appeared to be in English or Spanish, when it was supposed to be in another language, we set a flag called "quarantine" that prevented it from being released as part of the public dataset. Similarly, text strings that contain digits instead of spelled out numbers are flagged, such that they are not included by default.

The records output from the dataset are designed to be consistent with other speech recognition datasets on Hugging Face, with \texttt{file}, \texttt{audio}, and \texttt{text} fields. A field for \texttt{credits} was to comply with the Creative Commons attribution sharing requirements, and a field for \texttt{license} was added to identify subsets of the dataset that are licensed under specific Creative Commons licenses (such as cc-by-nc-nd and cc-by-sa). \texttt{book}, \texttt{instance}, and \texttt{original\_lang\_tag} provide a means of matching up content with other Bloom Library datasets. The training, test, and validation splits for each language followed the same methodology as that described for \texttt{bloom-captioning}. 

The \texttt{bloom-speech} dataset is listed in the Hugging Face Hub as \textit{bloom-speech}.\footnote{\url{https://huggingface.co/datasets/sil-ai/bloom-speech}}

\section{Dataset Analysis}

The final statistics per language family for all the datasets are presented in Table~\ref{tab:languages}. In total, the datasets include 11,407 text stories, 112,080 image-caption pairs, 25,680 audio files, and 2,873 minutes of audio across 363 languages and 32 language families. A full list of included languages is provided in the Appendix. 

\begin{table*}[]
\centering
\begin{tabular}{lrrrrr}
\toprule
\textbf{Language family} &  \textbf{Languages} &  \textbf{Stories} & \textbf{Audio files} &  \textbf{Audio minutes} &  \textbf{Image-caption pairs} \\ \hline
\midrule
Afro-Asiatic         &         19 &      304 &         799 &            111 &                 2329 \\
Algic                &          3 &        6 &          85 &             36 &                   80 \\
Austro-Asiatic       &         17 &      195 &           0 &              0 &                 2219 \\
Austronesian         &         59 &     1560 &        1660 &            160 &                12593 \\
Chibchan             &          1 &        1 &           0 &              0 &                    7 \\
Chocoan              &          1 &        0 &          19 &              1 &                    0 \\
Creole               &          7 &      532 &        1280 &            137 &                 5209 \\
Dravidian            &          8 &       56 &         818 &            251 &                  410 \\
Eyak-Athabaskan      &          1 &        1 &          14 &             13 &                   13 \\
Hmong-Mien           &          1 &       17 &           0 &              0 &                  212 \\
Indo-European        &         43 &     5961 &        7399 &            829 &                60735 \\
Japonic              &          1 &        1 &           0 &              0 &                    2 \\
Koreanic             &          1 &      132 &           0 &              0 &                 2773 \\
Kra-Dai              &          6 &      322 &           0 &              0 &                 3229 \\
Maipurean            &          2 &        3 &           0 &              0 &                   31 \\
Mayan                &          6 &      438 &        7212 &            641 &                 4556 \\
Niger-Congo          &        101 &      409 &        2704 &            402 &                 3703 \\
Nilo-Saharan         &         13 &       74 &          15 &              1 &                  871 \\
North Bougainville   &          1 &        1 &           0 &              0 &                    9 \\
Otomanguean          &          6 &       28 &          89 &              9 &                  245 \\
Panoan               &          1 &       13 &           0 &              0 &                  103 \\
Quechuan             &          8 &       18 &           0 &              0 &                  154 \\
Ramu-Lower Sepik     &          1 &        1 &           0 &              0 &                    7 \\
Sepik                &          4 &       16 &         420 &             37 &                  194 \\
Sino-Tibetan         &         25 &      768 &        2234 &            195 &                 6579 \\
South Bougainville   &          3 &        7 &          63 &              6 &                   53 \\
South-Central Papuan &          1 &        7 &         275 &             15 &                  113 \\
Tequistlatecan       &          2 &        2 &         505 &             16 &                  109 \\
Torricelli           &          1 &        1 &           0 &              0 &                   11 \\
Trans-New Guinea     &         15 &      103 &          89 &             13 &                  930 \\
Turkic               &          2 &      411 &           0 &              0 &                 4332 \\
Uto-Aztecan          &          3 &       19 &           0 &              0 &                  269 \\
\hline
\midrule
\textbf{Total} & \textbf{363} & \textbf{11407} & \textbf{25680} & \textbf{2873} & \textbf{112080} \\
\bottomrule
\end{tabular}
\caption{\label{tab:languages} Language coverage and dataset statistics by language family. In total, \languagecountacrossalldatasets{} language families are represented.}
\end{table*}

Our \texttt{bloom-vist}, \texttt{bloom-lm}, and \texttt{bloom-captioning} datasets include data from 351 of these languages across 31 language families. There is a mean of 32 stories and 319 image-caption pairs and median of 2 stories and 22 image-caption pairs per language. Our \texttt{bloom-speech} dataset comprises 428 hours of total audio in \languagecountinbloomspeech{} languages. There is a mean of 458 and median of 138 records per language. 18 language families are represented, with a mean of 159 minutes and median of 31 minutes of audio per language family. A full breakdown of the composition of the dataset is available in the Appendix (Table~\ref{tab:speechsplits}).  

Notably for \texttt{bloom-speech}, among some of the languages of wider use in the dataset (e.g. French, Spanish, and English) the accent of the speaker is often not the same as in other public datasets. In Common Voice, a large majority of French data is labelled as coming from speakers in France. Most of the French data in \texttt{bloom-speech} is from Francophone Africa. The Spanish is primarily from Central America, and the English includes many places where English is a language of commerce. 

\section{Baseline Experiments}

To establish some of the first known baselines in languages included in Bloom Library and to assess the difficulty of language modeling, image captioning, and automatic speech recognition with the Bloom Library datasets, we trained baseline models for 44 included languages. The summarized results are included in Table~\ref{tab:results}.

\begin{table*}[]
\centering
\begin{tabular}{l|lll|llll|llll}
\multicolumn{1}{c|}{\textbf{}} & \multicolumn{3}{c|}{\textbf{Language Modeling}}                    & \multicolumn{4}{c|}{\textbf{Image Captioning}}  & \multicolumn{4}{c}{\textbf{Speech Recognition}} \\ \hline
\textbf{ISO}   & \textbf{Stories} & \textbf{PPL} & \textbf{ACC} & \textbf{Pairs} & \textbf{BLEU} & \textbf{chrF2} & \textbf{TER} & \textbf{Files} & \textbf{Min.} & \textbf{WER} & \textbf{CER} \\ \hline
     ahk &      101 &   3.06 &  0.75 &                  907 &   0.7 &  16.6 &  115.1 &            0 &              0 &     - &     - \\
     awa &      163 &  10.54 &  0.54 &                 1200 &   0.1 &   5.4 &  103.4 &            0 &              0 &     - &     - \\
     bam &        4 &      - &     - &                   86 &     - &     - &      - &          303 &             52 &  1.06\textsuperscript{*} &  1.04\textsuperscript{*} \\
     ben &      251 &    4.5 &  0.66 &                 2235 &   7.1 &  13.9 &  112.8 &            0 &              0 &     - &     - \\
     bho &      173 &   8.56 &  0.56 &                 1172 &   0.1 &   6.1 &  150.7 &            0 &              0 &     - &     - \\
     boz &        5 &      - &     - &                  102 &     - &     - &      - &          529 &             53 &  0.35\textsuperscript{*} &  0.09\textsuperscript{*} \\
     bzi &       66 &      - &     - &                  497 &     - &     - &      - &         1570 &            123 &  0.11 &  0.02 \\
     cak &       67 &  13.41 &  0.55 &                  817 &   8.6 &  19.2 &  138.6 &         1154 &            123 &  0.21\textsuperscript{*} &  0.04\textsuperscript{*} \\
     ceb &      418 &  16.78 &  0.51 &                 2953 &     - &     - &      - &          670 &             61 &  0.18\textsuperscript{\textdagger} &  0.04\textsuperscript{\textdagger} \\
     cgc &      197 &  33.59 &  0.44 &                 1638 &   0.0 &   6.7 &  103.1 &            0 &              0 &     - &     - \\
     chd &        1 &      - &     - &                   84 &     - &     - &      - &          305 &              7 &  1.06\textsuperscript{*} &  0.57\textsuperscript{*} \\
     dty &      172 &  10.84 &  0.52 &                 1310 &   0.8 &   7.6 &  114.0 &            0 &              0 &     - &     - \\
     eng &     2633 &   6.96 &   0.6 &                28618 &  13.9 &  25.4 &  126.0 &         4646 &            525 &  0.27\textsuperscript{\textdagger} &   0.1\textsuperscript{\textdagger} \\
     fas &      129 &  17.17 &  0.45 &                  631 &   0.2 &   6.4 &  140.9 &            0 &              0 &     - &     - \\
     fra &      403 &   5.62 &  0.63 &                 5278 &  14.8 &  25.0 &  113.0 &          360 &             86 &  0.29\textsuperscript{\textdagger} &  0.09\textsuperscript{\textdagger} \\
     hat &      260 &  14.85 &  0.51 &                 2411 &   1.6 &  15.7 &  111.3 &            0 &              0 &     - &     - \\
     hau &      256 &  15.04 &  0.54 &                 1865 &  19.6 &  32.8 &   93.2 &            0 &              0 &     - &     - \\
     hbb &       27 &      - &     - &                  273 &     - &     - &      - &          675 &             95 &  0.27 &  0.06 \\
     ind &      259 &    8.3 &   0.6 &                 2177 &   3.8 &  16.6 &  110.0 &           14 &              1 &     - &     - \\
     jra &      139 &   5.26 &  0.67 &                 1423 &   0.2 &   6.3 &  132.1 &          303 &             33 &  0.12 &  0.03 \\
     kak &      195 &  17.85 &  0.52 &                 1416 &   0.2 &   7.9 &  107.1 &            0 &              0 &     - &     - \\
     kan &       21 &      - &     - &                  168 &     - &     - &      - &          374 &            119 &  0.47\textsuperscript{\textdagger} &  0.13\textsuperscript{\textdagger} \\
     kek &       36 &  13.09 &  0.59 &                  621 &   0.4 &  11.9 &  145.7 &         1915 &            168 &  0.51\textsuperscript{*} &  0.13\textsuperscript{*} \\
     kir &      382 &   5.98 &  0.62 &                 4026 &  20.9 &  26.0 &  126.7 &            0 &              0 &     - &     - \\
     kjb &      102 &  11.49 &  0.56 &                  984 &   0.2 &  13.4 &  180.4 &          911 &             61 &  0.33\textsuperscript{*} &  0.11\textsuperscript{*} \\
     kor &      132 &   9.95 &  0.56 &                 2773 &  10.1 &  14.5 &  104.1 &            0 &              0 &     - &     - \\
     mai &      180 &   7.78 &  0.59 &                 1211 &   0.8 &  10.0 &  103.1 &           11 &              3 &     - &     - \\

     mai &      180 &   7.78 &  0.59 &                 1211 &   0.8 &  10.0 &  103.1 &           11 &              3 &     - &     - \\
     mam &      134 &   9.41 &  0.58 &                 1317 &   8.0 &  17.3 &  189.1 &         1514 &            129 &  0.32\textsuperscript{*} &  0.07\textsuperscript{*} \\
     mhx &       98 &   7.02 &   0.6 &                  945 &   3.5 &  14.1 &  175.1 &            0 &              0 &     - &     - \\
     mya &       38 &      - &     - &                  421 &     - &     - &      - &          421 &             60 &   0.8\textsuperscript{\textdagger} &  0.09\textsuperscript{\textdagger} \\
     myk &       34 &      - &     - &                  341 &     - &     - &      - &          799 &            113 &  0.21\textsuperscript{*} &  0.05\textsuperscript{*} \\
     nep &      200 &   5.84 &  0.63 &                 1507 &   0.8 &   7.9 &  106.2 &            0 &              0 &     - &     - \\
     new &      177 &   7.41 &  0.57 &                 1225 &   0.0 &   6.4 &  108.7 &            0 &              0 &     - &     - \\
     por &      163 &   6.92 &   0.6 &                 3101 &  11.9 &  22.9 &  117.5 &           34 &              3 &     - &     - \\
     quc &       99 &  21.02 &  0.48 &                  817 &   3.4 &  12.4 &  179.2 &         1677 &            154 &  0.31\textsuperscript{*} &  0.08\textsuperscript{*} \\
     rus &      353 &   6.62 &   0.6 &                 3933 &  13.3 &  24.7 &  187.9 &            0 &              0 &     - &     - \\
     sdk &       11 &      - &     - &                  153 &     - &     - &      - &          412 &             36 &  0.28 &  0.06 \\
     snk &       35 &      - &     - &                  356 &     - &     - &      - &          662 &             88 &   0.3 &  0.07 \\
     spa &      528 &   7.24 &  0.59 &                 6111 &  10.2 &  18.7 &  131.7 &         2073 &            148 &  0.24\textsuperscript{\textdagger} &  0.08\textsuperscript{\textdagger} \\
     stk &        7 &      - &     - &                  113 &     - &     - &      - &          275 &             15 &  0.51\textsuperscript{*} &  0.16\textsuperscript{*} \\
     tgl &        0 &      - &     - &                    0 &     - &     - &      - &          450 &             38 &  0.18\textsuperscript{\textdagger} &  0.05\textsuperscript{\textdagger} \\
     tha &      285 &   3.81 &  0.67 &                 3023 &  31.2 &  34.0 &  101.5 &            0 &              0 &     - &     - \\
     thl &      185 &   9.16 &  0.55 &                 1464 &   5.1 &  10.2 &  135.0 &            0 &              0 &     - &     - \\
     tpi &      201 &   4.22 &  0.75 &                 2162 &  29.2 &  38.8 &  102.1 &         1234 &            131 &  0.09\textsuperscript{\textdagger} &  0.02\textsuperscript{\textdagger} \\
     \hline
\end{tabular}
\caption{\label{tab:results} Our complete baseline results for all tasks. \\
\textsuperscript{*}Languages from language families not represented at all in XLS-R \\
\textsuperscript{\textdagger}Languages explicitly included in the training for XLS-R, which we fine-tuned for our baselines. We may expect these scores to benefit compared to languages which were not.}
\end{table*}

\subsection{Language Modeling}

We used \texttt{bloom-lm} to fine-tune the DistilBERT base multilingual cased model~\cite{Sanh2019DistilBERTAD} (available on the Hugging Face hub\footnote{\url{https://huggingface.co/distilbert-base-multilingual-cased}}) for 32 languages. These particular languages were chosen because they each had more than 500 stories in the Bloom Library. 

We fine-tuned these models on the training split of \texttt{bloom-lm} and tested on the test set for the masked language modeling task. Training was implemented using the Hugging Face Trainer API from the Python \texttt{transformers} package version 4.20.1~\cite{wolf-etal-2020-transformers}, with a training and evaluation batch size of 8 and a random seed of 1022. The default Trainer API configurations were used for all other hyperparameters and training arguments. The models each ran for 3 epochs on P100 GPUs. 

Perplexity and accuracy were used as evaluation metrics for the masked language modeling task. The full results can be seen in Table~\ref{tab:results}. The mean perplexity for all the languages was 10.29, with a maximum of 33.59 (for Kagayanen [cgc]) and minimum of 3.06 (for Akha [ahk]). For reference, the RoBERTa base model trained over BOOKCORPUS and WIKIPEDIA achieves a perplexity of 3.68~\cite{Liu2019RoBERTaAR}. Thus, \texttt{bloom-lm} seems to show promise for kickstarting language modeling tasks in many new languages, especially when fine-tuning from existing multilingual language models.  

\subsection{Image Captioning}

We used \texttt{bloom-captioning} to train image captioning models inspired by \citet{10.5555/3045118.3045336} for 31 languages. For each language, we first downloaded the image-caption pairs and performed some pre-processing on the data. For the images, we resized each image to 299x299 pixels and normalized the images so that they contained pixels in the range of -1 to 1. We then extracted image features for each of the images using a version of InceptionV3~\cite{7780677} pretrained on Imagenet and available in TensorFlow version 2.8.0. For the captions, we encoded the text into integer sequences with the \texttt{TextVectorization} layer within TensorFlow Keras, keeping a vocabulary of the top 5,000 words.  

The image captioning model used a Convolutional encoder network (CNN) followed by an Recurrent decoder network (RNN). The shape of the features from InceptionV3 was 2,048, and we used an embedding dimension of 256 along with a hidden attention layer (Bahdanau Attention) having a dimension of (batch size, 64, 512) in the RNN decoder. Each image captioning model was trained for 50 epochs on A100 GPUs using a random seed of 1022, the default settings of the TensorFlow Adam optimizer, and Sparse Categorical Cross-entropy loss. 
BLEU score, chrF2, and Translation Error Rate (TER) were used as evaluation metrics for the image captioning task. The full results can be seen in Table~\ref{tab:results}. The mean BLEU score for all languages was 7.12, with a maximum of 31.2 (for Thai [tha]) and a minimum of 0.0 (for Kagayanen [cgc] and Newar [new]). For reference, the state-of-the-art result on the COCO captions dataset~\cite{https://doi.org/10.48550/arxiv.1504.00325} (at the time of this paper was drafted) is a BLEU score of 44.9~\cite{https://doi.org/10.48550/arxiv.2202.03052}. 

Thai, which has the best captioning performance, also shows good results in language modeling having a perplexity of 3.81. This trend is not generally true, however, with Akha achieving a BLEU score of 0.7 despite having the best language modeling performance. This is likely due to the number of stories and captions available in each language (3k+ image-caption pairs for Thai and only around 900 for Akha) and the diversity of those stories and captions.

\subsection{Speech Recognition}

We used \texttt{bloom-speech} to fine-tune Wav2Vec2 XLS-R model~\cite{xls-r} (available on the Hugging Face hub) for 23 languages --- any language with at least 275 audio files. Training was implemented using the Hugging Face Trainer API from the Python \texttt{transformers} package installed from source at a particular GitHub commit ID\footnote{7cf52a49dee661f6adb7847991c6a84925999f5d}~\cite{wolf-etal-2020-transformers}.

For the text strings, we normalized whitespace characters (eliminating carriage returns, non-breaking spaces, etc.) and removed special characters. For a full list of the special characters, see the source code.\footnote{\url{https://github.com/sil-ai/bloom-speech-training}}
No adjustments were needed for audio processing, as the Hugging Face code and libraries standardize the audio to 16khz sampling rate, as required by Wav2Vec2-based models.

All languages were trained with the same parameter settings. Except otherwise noted, all parameters and configuration were taken from the "Single GPU CTC" settings in the Hugging Face Transformers speech recognition example~\cite{hf-speech-recognition}. Early stopping was used with a patience of three, tracking the Word Error Rate (WER) metric. \texttt{max\_duration\_in\_seconds} was set to 25, approximately the maximum that the 20GB GPU partitions we were using could support. \texttt{eval\_steps} and \texttt{warmup\_steps} were set to 250, with \texttt{save\_steps} set to 500. Because of filtering for max duration and special characters, the size of the train, validation, and test sets is reduced in the experiment. For purposes of reproducibility and comparison, we provide the before and after split sizes in the Appendix, Table~\ref{tab:speechsplits}. 

The complete baseline data for all 23 languages is available in Table~\ref{tab:results}. The mean WER across all sets is 0.37 and for all sets with over one hour of training audio, the mean WER is 0.25. Tok Pisin~[tpi], Bisu~[bzi], and Jarai~[jra] achieve the top three score with 0.09, 0.11, and 0.12 WER, in a range that we consider among the state-of-the-art for languages in low resource settings. Tok Pisin is one of the languages explicitly included in the training data for XLS-R, while data from Bisu and Jarai were not included (but may benefit from adjacent languages). Of the subset of languages not found in XLS-R, Kaqchikel~[cak] (of the Mayan language family) achieves the best score with WER of 0.21, which may be practically useful for certain implementations. 

We believe the case of the Bisu language is particularly worth noting as the language has a population of as few as 700 users~\cite{david_m__eberhard_ethnologue_2021}. The efforts of the Bisu language community to preserve their language~\cite{bisu}, particularly through the creation of Bloom Books, and the work that has gone into making large multilingual speech models like XLS-R puts advanced language technology within reach using our datasets. 

\section{Conclusion}

We present 4 datasets processed out of the Bloom Library books. These datasets represent a massive increase in language and modality coverage for the tasks of language modeling, image captioning, visual storytelling, and speech recognition. In total, the datasets cover 363 languages across 32 language families. 

We demonstrated an ability to make use of the data as-is, and we foresee this dataset being used by researchers to fine-tune multilingual models (as recommended by, e.g., \citet{xls-r}), benchmark zero-shot performance for linguistically diverse languages, or train new models from scratch in combination with other datasets. 

In future work, we would like to continue to improve the size and quality of the dataset. We would also like to explore and understand baseline performance in the various tasks. Some of the baseline performance numbers are quite low compared to numbers on established datasets, and it would be interesting to further understand the characteristics of the Bloom data that make it challenging for various tasks (e.g., lexical diversity). Finally, we would like to prepare aligned mutlilingual versions of the dataset that would be immediately useful for cross-lingual and multilingual tasks that require parallel corpora. 

\section{Limitations}
\label{limits}

\subsection{Data quality}
The Bloom Library is populated primarily by community submissions. This increases the linguistic and topic diversity of the set, but it also leads to issues with user-submitted books that have inconsistent metadata, varying quality, images not matching captions, etc. While many of these issues can be detected during the parsing process (e.g. if the number of images does not match the number of captions), it is likely that some issues persist and will need to be addressed in future releases. 

\subsection{Sources of bias}
Of the religious books included in the datasets, there is a bias towards Christian books and Bible stories. However, the datasets also covers a wide variety of non-religious topics of interest to local language communities and fitting their local language ecology. Some of the non-religious topics included in the datasets are STEM (307 books), COVID-19 (370 books), and Agriculture (87 books). 

Generally, there may be few individuals or organizations producing the underlying content for a given language. Bias should be expected and appropriate steps should be taken to evaluate and counteract biases depending on how the data is used.

\subsection{Cross-lingual alignment}
While our datasets cover many languages, all of the data (for any one particular language) should be considered to be monolingual and unaligned. Bloom books are, by their nature, community-submitted books. The format, count, or even order of image-caption pairs is not guaranteed to match across books within the same translation "lineage." Some translations may, for example, lack captions for certain images. In this release of the datasets, we therefore release each dataset in monolingual form without attempting to align for tasks such as machine translation. 

\subsection{Audio and speakers}
For many languages, we have few audio files. Practically, some of these may only serve as a supplement to another dataset by providing a separate domain test set. We have not assessed the variety of speakers in any language, but assume that in some languages there will only be one or a few speakers. In training speech recognition systems, this may create problems in recognizing different speakers (also see bias in \ref{ethics}). Fine-tuning from a large multilingual speech language model, as we have done, may offset some of these concerns. This may also make our metrics (WER and CER) less comparable with other datasets having more variety. While comparison between different datasets/domains is always a concern with these metrics, it may be amplified here. Our models are intended only as baselines and have not been customized to the particularities of each individual language.

\subsection{Reproducibility}

All of the source code for dataset preparation and baseline model training/evaluation is included in this GitHub repository.\footnote{\url{https://github.com/sil-ai/bloom-parsing}}. All of the data used for baseline experiments has been released on the Hugging Face hub.\footnote{\url{https://huggingface.co/datasets?search=sil-ai+bloom}}

\section{Ethical considerations}
\label{ethics}

Books in the released datasets are restricted to those available on the Bloom Library under a Creative Commons (CC) license. The original creators of the material chose these CC licenses, and we received expressed permission for this use and access from Bloom Library team consistent with their terms of use. The original creators were aware at the time of creating the materials that their material would be published publicly on the Internet, and are presumed to be aware that this data would be widely available for uses beyond those originally envisioned.

In this paper we aim to demonstrate the technical feasibility of applications of this dataset to creating NLP tools in local languages. This should not be taken to suggest that we recommend using these tools in local contexts. The normative ethics around creating a tool for African French speakers, for example, do not necessarily apply for local languages like Bisu. We would recommend a consideration of the language ecology~\cite{bird-2022-local} and designing tools for "conviviality"~\cite{conviviality}, with the goal of sustaining language use~\cite{sustaininglanguage}. That is, we urge the reader to consider the impact of the tools they are creating on local language communities, enhance the agency of the users and community who use (or do not use) the tool, and support the flourishing of the community through the use of its own language. 

We assess the ecological impact of our dataset creation and model training to be minimal. The dataset creation made minimal use of energy hungry GPUs, and was not measured. Total training time for \texttt{bloom-speech} based models was 86 hours, an average of 3 hours and 45 minutes per model trained. Each training was conducted on a 20G or 40G partition of an NVIDIA A100. Each of the image captioning baseline models and language modeling baseline models required even less training time than \texttt{bloom-speech}.

In terms of equitable access, the hardware needed to replicate the experiments is available through cloud services at a moderate price.  The data is being released and will be made publicly available under CC licenses for primarily non-commercial use. 

\section*{Acknowledgments}

The Masakhane community has been a great source of help, feedback, and inspiration. Colin Leong would like to thank the University of Dayton Research Institute for their support and encouragement, as well as Dr. Vijayan Asari and the University of Dayton Vision Lab for support and guidance. Further, the authors appreciate important feedback on the paper draft and draft version of the datasets from Sebastian Ruder, Genie Razumovskaia, Paul Frank, and Gary Simons.

\bibliography{anthology,custom}
\bibliographystyle{acl_natbib}

\appendix

\section{Appendix}
\label{sec:appendix}

The full list of languages included in this initial release of our datasets is as follows:

Ghotuo[aaa]; Ayta, Ambala[abc]; Dangme[ada]; Adangbe[adq]; Akeu[aeu]; Afrikaans[afr]; Aghem[agq]; Esimbi[ags]; Akha[ahk]; Arosi[aia]; Amri Karbi[ajz]; Akan[aka]; Yanesha’[ame]; Amharic[amh]; Alamblak[amp]; Amuzgo, Guerrero[amu]; Obolo[ann]; Athpariya[aph]; Awadhi[awa]; Awa[awb]; Nahuatl, Western Durango[azn]; Awing[azo]; Tuki[bag]; Bamanankan[bam]; Bambili-Bambui[baw]; Bamun[bax]; Babanki[bbk]; Balochi, Southern[bcc]; Bamenyam[bce]; Iceve-Maci[bec]; Benabena[bef]; Bengali[ben]; Bafut[bfd]; Mmen[bfm]; Bunak[bfn]; Bangandu[bgf]; Bhojpuri[bho]; Buwal[bhs]; Bislama[bis]; Banjar[bjn]; Binumarien[bjr]; Baka[bkc]; Bakoko[bkh]; Kom[bkm]; Baikeno[bkx]; Aweer[bob]; Tibetan, Central[bod]; Bozo, Tieyaxo[boz]; Wumboko[bqm]; Braj Bhasha[bra]; Lave[brb]; Mokpwe[bri]; Bru, Western[brv]; Akoose[bss]; Ntcham[bud]; Terei[buo]; Bafaw-Balong[bwt]; Bunu, Bu-Nao[bwx]; Tairaha[bxa]; Bukusu[bxk]; Batak[bya]; Bozo, Jenaama[bze]; Bisu[bzi]; Kaqchikel[cak]; Kakataibo-Kashibo[cbr]; Cebuano[ceb]; Kagayanen[cgc]; Chontal, Highland Oaxaca[chd]; Dene[chp]; Cimbrian[cim]; Kurdish, Central[ckb]; Chontal, Lowland Oaxaca[clo]; Chinese, Mandarin[cmn]; Chinese, Mandarin[cmn]; Mnong, Central[cmo]; Cree, Swampy[csw]; Gichuka[cuh]; Cuvok[cuv]; Dagbani[dag]; Fataluku[ddg]; Dedua[ded]; German, Standard[deu]; Chidigo[dig]; Zarma[dje]; Kinabatangan, Upper[dmg]; Dani, Western[dnw]; Kadazan Dusun[dtp]; Lotud[dtr]; Dotyali[dty]; Chiduruma[dug]; Elip[ekm]; Markweeta[enb]; En[enc]; English[eng]; Ewondo[ewo]; Filipino[fil]; Fali[fli]; Fon[fon]; French[fra]; Fulfulde, Adamawa[fub]; Fulfulde, Western Niger[fuh]; Galolen[gal]; Gadaba, Bodo[gbj]; Gavar[gou]; German, Swiss[gsw]; Wayuu[guc]; Gujarati[guj]; Ekegusii[guz]; Gawri[gwc]; Hakö[hao]; Haitian Creole[hat]; Hausa[hau]; Nya Huba[hbb]; Kamwe[hig]; Hiligaynon[hil]; Hindi[hin]; Halia[hla]; Mina[hna]; Hre[hre]; Haroi[hro]; Idaté[idt]; Ilocano[ilo]; Indonesian[ind]; Inoke-Yate[ino]; Isu[isu]; Italian[ita]; Ngomba[jgo]; Mixtec, Western Juxtlahuaca[jmx]; Japanese[jpn]; Jarai[jra]; Kalanguya[kak]; Kamba[kam]; Kannada[kan]; Kamano[kbq]; Ap Ma[kbx]; Kanuri, Manga[kby]; Kanuri, Manga[kby]; Q’eqchi’[kek]; Kenyang[ken]; Lü[khb]; Khmer[khm]; Gikuyu[kik]; Kinyarwanda[kin]; Kyrgyz[kir]; Q’anjob’al[kjb]; Kâte[kmg]; Kurdish, Northern[kmr]; Kamasau[kms]; Kanite[kmu]; Korean[kor]; Kimaragang[kqr]; Krung[krr]; Karen, S’gaw[ksw]; Lahta[kvt]; Kwaio[kwd]; Kwakum[kwu]; Khirwar[kwx]; Koli, Wadiyari[kxp]; Kenga[kyq]; Lango[laj]; Laru[lan]; Lao[lao]; Lohorung[lbr]; Lefa[lfa]; Lugbara[lgg]; Lengo[lgr]; Lhomi[lhm]; Lahu[lhu]; Lukabaras[lkb]; Lole[llg]; Limbum[lmp]; Lamnso’[lns]; Narim[loh]; Lacid[lsi]; Lutachoni[lts]; Ganda[lug]; Lawa, Eastern[lwl]; Maithili[mai]; Malayalam[mal]; Mam[mam]; Marathi[mar]; Mandar[mdr]; Matal[mfh]; Mefele[mfj]; Mpumpong[mgg]; Mambae[mgm]; Meta’[mgo]; Malila[mgq]; Lhao Vo[mhx]; Mixtec, Ayutla[miy]; Makasae[mkz]; Manambu[mle]; Kiwilwana[mlk]; Moloko[mlw]; Mmaala[mmu]; Naba[mne]; Mundani[mnf]; Mon[mnw]; Barí[mot]; Mamasa[mqj]; Cheke Holo[mrn]; Mandaya[mry]; Masbatenyo[msb]; Muthuvan[muv]; Marwari[mve]; Mada[mxu]; Burmese[mya]; Sénoufo, Mamara[myk]; Masaaba[myx]; Mumuye[mzm]; Naasioi[nas]; Sibe[nco]; Newar[new]; Ngemba[nge]; Ngwo[ngn]; Nahuatl, Isthmus-Mecayapan[nhx]; Njyem[njy]; Ngombale[nla]; Dutch[nld]; Nahuatl, Orizaba[nlv]; Thai, Northern[nod]; Nepali[npi]; Naskapi[nsk]; Nehan[nsn]; Sotho, Northern[nso]; Naga, Tangshang[nst]; Nyole[nuj]; Ngwe[nwe]; Tanna, Southwest[nwi]; Nauete[nxa]; Nuaulu, South[nxl]; Chichewa[nya]; Nyoro[nyo]; Nyungwe[nyu]; Mbembe, Tigon[nza]; Oadki[odk]; Oji-Cree[ojs]; Okiek[oki]; Tairora, South[omw]; Odia[ory]; Koonzime[ozm]; Pagibete[pae]; Pangasinan[pag]; Punjabi, Eastern[pan]; Pashto, Southern[pbt]; Palaung, Ruching[pce]; Paniya[pcg]; Kayan[pdu]; Indonesian, Peranakan[pea]; Persian, Iranian[pes]; Petats[pex]; Pijin[pis]; Kipfokomo[pkb]; Pamona[pmf]; Pana[pnz]; Portuguese[por]; Gapapaiwa[pwg]; Quechua, Huallaga[qub]; K’iche’[quc]; Quechua, Lambayeque[quf]; Quechua, Cusco[quz]; Quechua, Eastern Apurímac[qve]; Quechua, Huamalíes-Dos de Mayo Huánuco[qvh]; Quechua, Margos-Yarowilca-Lauricocha[qvm]; Quichua, Napo[qvo]; Quechua, Panao[qxh]; Rendille[rel]; Ranglong[rnl]; Romanian[ron]; Rotokas[roo]; Rusyn[rue]; Roviana[rug]; Russian[rus]; Sanskrit[san]; Samburu[saq]; Santhali[sat]; Sos Kundi[sdk]; Semai[sea]; Surigaonon[sgd]; Shan[shn]; Sama, Central[sml]; Soninke[snk]; Sangil[snl]; Somali[som]; Sotho, Southern[sot]; Swo[sox]; Spanish[spa]; Saposa[sps]; Waata[ssn]; Arammba[stk]; Swahili[swh]; Swahili[swh]; Suba[sxb]; Syuba[syw]; Tamang, Eastern[taj]; Tamil[tam]; Tiang[tbj]; Panchpargania[tdb]; Emberá-Tadó[tdc]; Tamang, Western[tdg]; Tetun Dili[tdt]; Teso[teo]; Tetun[tet]; Tajik[tgk]; Tagalog[tgl]; Thai[tha]; Tharu, Madhya Ksetriya[the]; Kitharaka[thk]; Tharu, Dangaura[thl]; Tha[thy]; Teop[tio]; Tukudede[tkd]; Lenakel[tnl]; Tanna, North[tnn]; Whitesands[tnp]; Tontemboan[tnt]; Toma[tod]; Tombulu[tom]; Tok Pisin[tpi]; Me’phaa, Tlacoapa[tpl]; Tampuan[tpu]; Tsamai[tsb]; Setswana[tsn]; Tsonga[tso]; Turkana[tuv]; Turka[tuz]; Taveta[tvs]; Tz’utujil[tzj]; Muduga[udg]; Mundari[unr]; Urdu[urd]; Uzbek, Northern[uzn]; Venda[ven]; Vietnamese[vie]; Vili[vif]; Waray-Waray[war]; Wa, Vo[wbm]; Wagdi[wbr]; Wambon[wms]; Comorian, Ndzwani[wni]; Wanukaka[wnk]; Watakataui[wtk]; Xhosa[xho]; Kagoro[xkg]; Mbudum[xmd]; Mengaka[xmg]; Malay, Manado[xmm]; Soga[xog]; Mixtec, Yoloxóchitl[xty]; Nugunu[yas]; Yangben[yav]; Yemba[ybb]; Yakkha[ybh]; Yamphu[ybi]; Yiddish, Eastern[ydd]; Yiddish, Eastern[ydd]; Ravula[yea]; Riang Lai[yin]; Yamap[ymp]; Zapotec, Mitla[zaw]; Malay[zlm]; Tokano[zuh]; Zulu[zul]

\begin{table*}[]
\centering
\begin{tabular}{l|lll}
\textbf{language} & \textbf{used/total train cuts} & \textbf{used/total validation cuts} & \textbf{used/total test cuts} \\
\hline
bam & 179 / 203	 & 43 / 50 & 45 / 50					      \\
boz & 425 / 427	 & 50 / 50 & 51 / 52					      \\
bzi & 1363 / 1363	 & 50 / 50 & 157 / 157					      \\
cak & 989 / 989	 & 50 / 50 & 115 / 115					      \\
ceb & 553 / 553	 & 50 / 50 & 67 / 67					      \\
chd & 205 / 205	 & 50 / 50 & 50 / 50					      \\
eng & 3979 / 4143	 & 46 / 48 & 445 / 455					      \\
fra & 208 / 261	 & 42 / 49 & 42 / 50					      \\
hbb & 546 / 558	 & 49 / 50 & 66 / 67					      \\
jra & 203 / 203	 & 49 / 50 & 50 / 50					      \\
kan & 232 / 281	 & 34 / 43 & 40 / 50					      \\
kek & 1675 / 1676	 & 49 / 49 & 189 / 190					      \\
kjb & 767 / 770	 & 50 / 50 & 91 / 91					      \\
mam & 1312 / 1313	 & 50 / 50 & 151 / 151					      \\
mya & 299 / 321	 & 49 / 50 & 48 / 50					      \\
myk & 635 / 669	 & 48 / 50 & 76 / 80					      \\
quc & 1449 / 1460	 & 50 / 50 & 166 / 167					      \\
sdk & 312 / 312	 & 50 / 50 & 50 / 50					      \\
snk & 517 / 546	 & 49 / 50 & 64 / 66					      \\
spa & 1807 / 1816	 & 50 / 50 & 207 / 207					      \\
stk & 180 / 180	 & 45 / 45 & 50 / 50					      \\
tgl & 352 / 352	 & 48 / 48 & 50 / 50					      \\
tpi & 1044 / 1061	 & 48 / 50 & 121 / 123 \\
\hline
\end{tabular}
\caption{\label{tab:speechsplits} \texttt{bloom-speech} comparison of cuts used in training versus total available. In cases where the total validation cuts is not 50, it indicates a file(s) which failed to download.}
\end{table*}

\end{document}